\documentclass[runningheads]{llncs}
\usepackage{graphicx}

\usepackage{tikz}
\usepackage{comment}
\usepackage{amsmath,amssymb} 
\usepackage{color}
\usepackage{xcolor}
\usepackage{algorithm}
\usepackage{algpseudocode}
\usepackage{tabularx, booktabs, makecell, caption}
\usepackage{multirow}
\usepackage{adjustbox}
\usepackage{threeparttable}
\usepackage{graphicx}
\usepackage[accsupp]{axessibility}  
\usepackage{hyperref}
\hypersetup{
    colorlinks=true,
}

\makeatletter
\renewcommand*{\@fnsymbol}[1]{\ensuremath{\ifcase#1\or *\or \dagger\or \ddagger\or
   \mathsection\or \mathparagraph\or \|\or **\or \dagger\dagger
   \or \ddagger\ddagger \else\@ctrerr\fi}}
\makeatother

\begin{document}
\pagestyle{headings}
\mainmatter
\def\ECCVSubNumber{6974}  

\title{$k$-SALSA: $k$-anonymous synthetic averaging of retinal images via local style alignment} 


\titlerunning{Synthetic Averaging of Retinal Images via Local Style Alignment}
%
\author{Minkyu Jeon\inst{1,2} \and
Hyeonjin Park\inst{5,}\thanks{This work was performed while the author was at Korea University}\and
Hyunwoo J. Kim\inst{2,}$^\dagger$ \and
Michael Morley\inst{3,4,}$^\dagger$ \and
Hyunghoon Cho\inst{1,}\thanks{Corresponding authors}}

\authorrunning{M. Jeon et al.}
%
\institute{Broad Institute of MIT and Harvard, Cambridge, MA, USA\\ 
\email{\{mjeon, hhcho\}@broadinstitute.org} \and
Korea University, Seoul, Republic of Korea\\
\email{hyunwoojkim@korea.ac.kr} \and
Harvard Medical School, Boston, MA, USA \\ \and
Ophthalmic Consultants of Boston, Boston, MA, USA \\
\email{mgmorley@eyeboston.com}\and
NAVER CLOVA, Seoul, Republic of Korea \\ \email{hyeonjin.park.ml@navercorp.com}}
\newcommand{\minkyu}{\textcolor[rgb]{1,0,1}}
\newcommand{\correct}{\textcolor[rgb]{0,0,1}}
\maketitle
\begin{abstract}
 The application of modern machine learning to retinal image analyses offers valuable insights into a broad range of human health conditions beyond ophthalmic diseases. Additionally, data sharing is key to fully realizing the potential of machine learning models by providing a rich and diverse collection of training data. However, the personally-identifying nature of retinal images, encompassing the unique vascular structure of each individual, often prevents this data from being shared openly. While prior works have explored image de-identification strategies based on synthetic averaging of images in other domains (e.g. facial images), existing techniques face difficulty in preserving both privacy and clinical utility in retinal images, as we demonstrate in our work. We therefore introduce $k$-SALSA, a generative adversarial network (GAN)-based framework for synthesizing retinal fundus images that summarize a given private dataset while satisfying the privacy notion of $k$-anonymity. $k$-SALSA brings together state-of-the-art techniques for training and inverting GANs to achieve practical performance on retinal images. Furthermore, $k$-SALSA leverages a new technique, called local style alignment, to generate a synthetic average that maximizes the retention of fine-grain visual patterns in the source images, thus improving the clinical utility of the generated images. On two benchmark datasets of diabetic retinopathy (EyePACS and APTOS), we demonstrate our improvement upon existing methods with respect to image fidelity, classification performance, and mitigation of membership inference attacks. Our work represents a step toward broader sharing of retinal images for scientific collaboration. Code is available at \url{https://github.com/hcholab/k-salsa}. \footnotemark[1]\footnotetext[1]{This paper has been accepted at ECCV 2022}

\end{abstract}
\section{Introduction}
\label{sec:introduction}
Retinal imaging is a fast, non-invasive, and cost-effective platform to study a range of systemic diseases, e.g. cardiovascular and neurological disorders~\cite{Wagner20}. Recent advances in machine learning (ML) are accelerating this transformation, equipping researchers with tools to recognize clinically relevant biomarkers across diverse imaging modalities, such as fundus imaging and optical coherence tomography (OCT). Studies have demonstrated the effectiveness of deep learning models in predicting clinical traits such as cardiovascular risk factors as well as other health-related information such as age, sex, and smoking status~\cite{Korot21,Poplin18,Wisely22}.

However, privacy concerns prevent the sharing of retinal images, presenting a hurdle for ML in ophthalmology~\cite{EyeNet21,Tom20}. 
Despite not being legally recognized as a biometric identifier in certain cases (e.g. HIPAA~\cite{Hipaa02}), retinal images are widely regarded as sensitive because they include individual-specific patterns like blood vessel structure~\cite{Marino06}.
Reflecting these concerns,
medical institutions have begun to refrain from using retinal images in grand rounds, lectures, and publications, leading to difficulties in research and education. We aim to tackle these challenges by transforming retinal images to protect privacy while preserving clinical utility.

To this end, a seminal work by Newton et al.~\cite{Newton05} on face de-identification introduced a class of techniques known as the ``$k$-Same’’ algorithms, wherein mutually disjoint clusters of $k$ images in the dataset are individually replaced with a single representative synthetic image that summarizes the visual characteristics of the images in each cluster. This naturally leads to a synthetic dataset satisfying the classical privacy notion of $k$-anonymity~\cite{Sweeney02}, which requires that each data instance in the released dataset cannot be distinguished among at least $k$ underlying individuals. $k$-anonymity has been widely considered in the medical literature as a meaningful privacy notion and has been used as a core principle in real-world systems and polices~\cite{Garfinkel15,Gkoulalas14,Jakob20}. Furthermore, recent successes of generative adversarial networks (GANs)~\cite{Goodfellow16,Gui21} in synthesizing realistic images in diverse domains suggest a promising approach for generating high-quality representatives of individual clusters by taking an average of images in the latent embedding space of a GAN, also known as the $k$-Same-Net algorithm~\cite{Meden18}. 

Despite the promise of these approaches, several key challenges remain in applying these methods to retinal images. First, while several works have explored the use of GANs to generate synthetic retinal images~\cite{Burlina22,Burlina19,Chen21,Coyner20,Niu19}, none to our knowledge have addressed the problem of effectively summarizing these images in the latent space, making the feasibility of this approach for retinal images an unknown. Next, the difficulty of capturing fine-grain visual patterns of retinal images (e.g. hemorrhages or lipid deposits) poses an additional challenge in preserving the clinical utility of these images. Finally, because $k$-anonymity does not directly imply privacy (individual images could potentially be inferred from the average), a direct evaluation of privacy offered by $k$-anonymity in the context of retinal images is needed before these tools may be used in practice~\cite{paul2021defending}. 

To address these challenges, we developed $k$-SALSA, an end-to-end
pipeline for synthesizing a $k$-anonymous dataset given a private dataset of retinal images.
We modernize the approach of $k$-Same-Net to use state-of-the-art techniques for training and inverting GANs.
This allows us to map the source images to an embedding space, ``average'' them, and generate a representative synthetic image.
We improve upon the existing methodology of taking the Euclidean average of embedding vectors by introducing a new technique called \emph{local style alignment}, which aims to maximize the retention of local texture information from the source images. This ensures that the output keeps clinically relevant features.

We evaluate our pipeline on two benchmark datasets (APTOS and EyePACS) and demonstrate the enhanced visual fidelity of the synthetic images generated by our approach with respect to the Fr\'{e}chet inception distance~\cite{Heusel17}, a standard quality metric for synthetic images.
We also show that the synthetic dataset generated by our approach enables accurate training of downstream classifiers for predicting varying degrees of diabetic retinopathy.

Lastly, we evaluate the privacy of our approach with respect to membership inference attacks (MIA), where an adversary tries to predict whether a given retinal image was part of the cluster represented by a specific synthetic image.
Our results show that synthetic images generated by $k$-SALSA provide strong mitigation of MIA, while prior $k$-Same approaches using pixel-wise or eigenvector-based averaging fail to do so, despite the fact that all of these approaches ostensibly satisfy $k$-anonymity.

\textbf{Summary of our contributions.} 
(1) We demonstrate the feasibility of GAN-based $k$-anonymization of retinal images. (2) We present a modernized $k$-Same algorithm using state-of-the-art GAN techniques, which are crucial for practical performance. (3) We introduce a novel technique---local style alignment---for generating a synthetic average with enhanced fidelity and downstream utility. (4) We perform comprehensive experiments on two datasets, evaluating the fidelity, utility, and privacy of our method compared to existing techniques.

\section{Related Work}
Traditional approaches for removing identifying features from private images (e.g., faces and medical images) involve direct manipulation of pixels, including masking, blurring, and pixelation~\cite{Ribaric15,Bischoff07,Milchenko13,Schimke11}.
However, these heuristics have been found to provide insufficient privacy protection~\cite{Abramian19,Ravindra21}.
In response, a \emph{learning}-based approach to de-identification has been increasingly studied~\cite{Der21}.
This is enabled by recent advances in GANs, which intuitively provide a more powerful approach to manipulate images according to their intrinsic manifold~\cite{Gui21}.

The existing literature on GAN-based transformation of images for privacy protection largely focuses on face de-identification.
A common approach for formalizing privacy protection in this problem is to combine multiple images to obtain $k$-anonymity through the $k$-Same framework~\cite{Jourabloo15,Meden18,Newton05}.
A key difficulty in these works has been generating high-quality images that capture useful information in the original image.
To this end, recent works have focused on developing techniques to disentangle and preserve non-identity attributes of the image, such as pose and facial expression~\cite{Jeong21,Jourabloo15,Maximov20,Wu19}.
However, these methods are not directly applicable to our setting given the unclear distinction between identity vs. non-identity features in retinal images beyond the blood vessel structure.
GAN-based approaches to generate images with differential privacy~\cite{Dwork14} have also been proposed~\cite{Long21,Xu19}, but current techniques lead to significant degradation of image quality (see supplement for an example application to retinal images).

Several recent works have successfully explored the use of GANs for generating realistic retinal images.
Niu et al.~\cite{Niu19} proposed a method to generate an image consistent with the given pathological descriptors.
Both Zhou et al.~\cite{Zhou20} and Chen et al.~\cite{Chen21} developed GAN models to synthesize retinal images conditioned on a semantic segmentation to improve disease classification performance.
Yu et al.~\cite{Yu19} introduced multi-channel GANs that improve the quality of generated retinal images by separately considering different elements of the image, including the blood vessels and the optic disc. 

\section{Method}
\subsection{Overview of $k$-SALSA}

\begin{figure*}
    \centering
    \includegraphics[width=.89\textwidth]{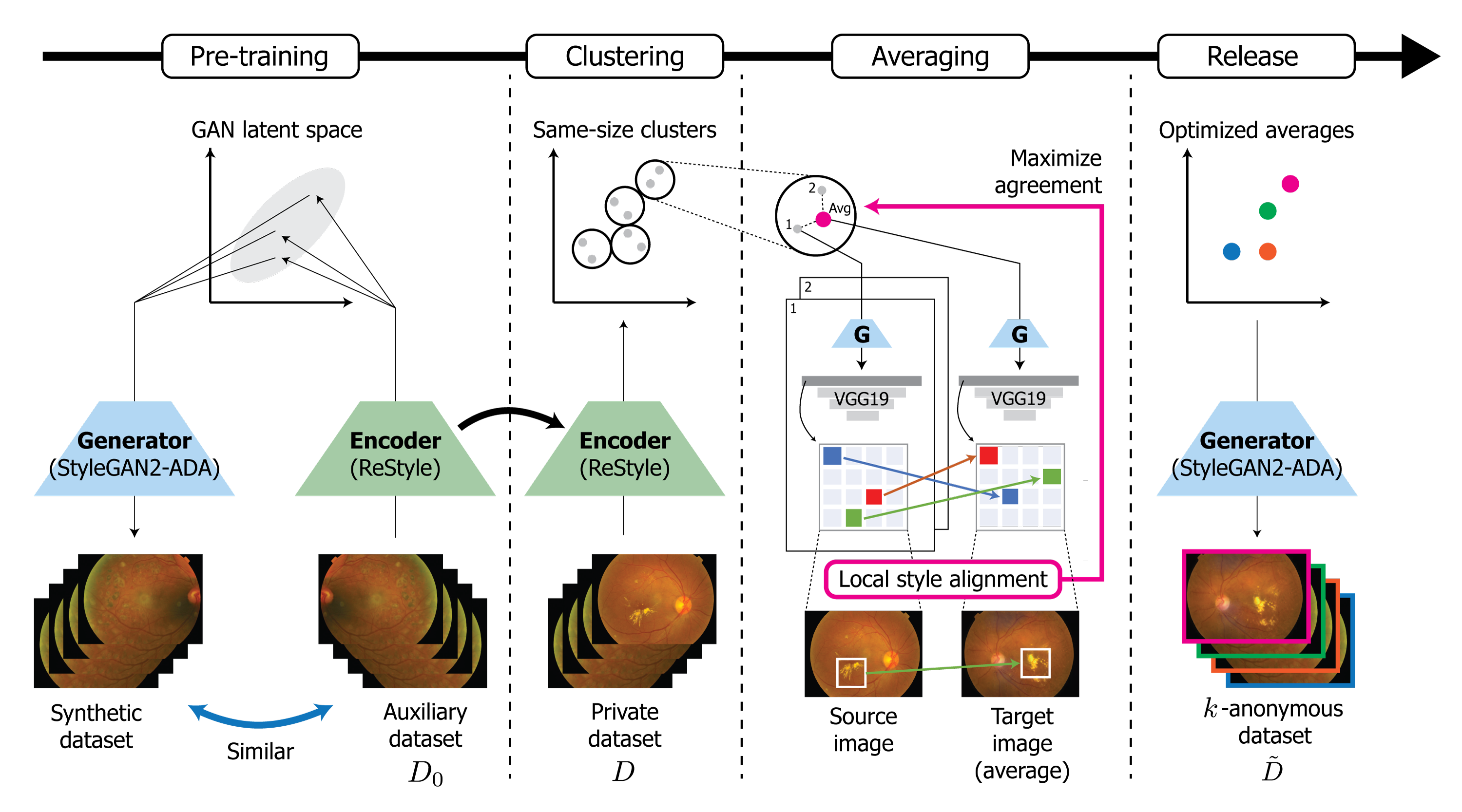}
    
    \caption{\textbf{Workflow of $k$-SALSA.} 
  GAN generator and inversion encoder are first trained to be used in subsequent steps. Same-size clustering groups images into groups of $k$, then the representative of each cluster is optimized via our local style alignment approach to preserve salient visual patterns. Images are synthesized from the optimized averages and released. Avg: average, G: generator
  }
    \label{fig:overview}
\end{figure*}

We consider a dataset $D$ of retinal fundus images that the user wishes to release in a privatized form. We assume that the user has access to an auxiliary dataset $D_0$ that can be used to pre-train the GAN components of $k$-SALSA.
Note that $D_0$ could simply be a publicly available dataset like the ones we used in our work or a subset of $D$.
This choice does not affect the  $k$-anonymity property of $k$-SALSA.
Given these datasets, $k$-SALSA proceeds in four steps:
\begin{enumerate}
    \item \textbf{Pre-training.} We first train a GAN on $D_0$.
Let $G(\cdot)$ be the trained generator, which maps a latent code $w \in \mathcal{W}$ to a synthetic image $G(w)$.
This step intuitively constructs the latent embedding space $\mathcal{W}$, which we will later rely on for the averaging operation.
Next, $k$-SALSA trains a GAN inversion model $E(\cdot)$ (can be viewed as an encoder), which maps an image $x\in \mathcal{X}$ to a particular latent code $E(x)\in \mathcal{W}$. Note that this inversion process is lossy in that the synthetic image $G(E(x))$ will only be approximately similar to the original image $x$, constrained by the limitations of encoder $E$ and generator $G$.  Together, these two functions approximate a bijection between the space of retinal images $\mathcal{X}$ and the latent embedding space $\mathcal{W}$.

    \item \textbf{Clustering.} Next, $k$-SALSA performs \emph{same-size clustering} of the target input dataset $D$ based on the inverted codes $E(x)$ for each $x\in D$, partitioning $D$ into groups of exactly $k$ similar images. Here $k$ is the user parameter determining the $k$-anonymity of the final output. If the size of $D$ is not divisible by $k$, one can disregard a small number of samples to resolve the issue, given that $k$ is typically small (e.g. 10). 
    \item \textbf{Averaging.} For each cluster, $k$-SALSA summarizes the $k$ source images as a single representative image via local style alignment---our new approach. This leads to $k$-anonymity, since each average image represents $k$ individuals as a whole and does not distinguish among them.
    \item \textbf{Release.}  Finally, $k$-SALSA constructs a $k$-anonymous synthetic dataset $\tilde{D}$ to release by associating each average image (one per cluster) with the aggregated labels of the $k$ images in the corresponding cluster (if labels were provided in the input dataset). This synthetic dataset can then be used for downstream analysis, such as training a classifier to predict the labels.
\end{enumerate}
A graphical illustration of our workflow is provided in Fig.~\ref{fig:overview}.

\subsection{Local Style Alignment: Our New Approach to Image Averaging}
\begin{algorithm}
\footnotesize
    \caption{$k$-SALSA}
    \label{alg:overall}
    \textbf{Input:} Private dataset $X=(x_1,\dots,x_n)$, auxiliary dataset $X_0$ for GAN model training, integer $k>1$ (assume $n = mk$ for integer $m$ without loss of generality), number of iterations $T$, loss ratio parameter $\lambda$ \\
    \textbf{Output:} Synthetic dataset $\tilde{X}$ of size $m$ with $k$-anonymity

    \begin{algorithmic}[1]
        \State Train a GAN generator $G$ and a GAN inversion encoder $E$ on $X_0$
        \State Obtain latent code $w_i = E(x_i)$ for each $i\in [n]$ and let $W=\{w_i\}_{i=1}^n$
        \State $(C_1,\dots,C_m)= \textsf{SameSizeClustering}(W, k)$  \Comment{$C_j\subset W$, $|C_j|=k$, $|C_j\cap C_{j'\neq j}|=0$, $\forall j$}
      
        \State Initialize $\tilde{X} = \emptyset$
        \For {each cluster $j\in [m]$}
        \State Let $C_j = (w_1',\dots,w_k')$, and $x_i'$ the original image of $w_i'$ for each $i$
        \State Compute $w_0 = \frac{1}{k} \sum_{i=1}^{k} w'_i$ and generate $x_0 = G(w_0)$
        \State Initialize $w_\text{avg}^{(0)} = w_0$
        \For {each iteration $t \in [T]$}
            \State Generate $x_\text{avg}^{(t-1)} = G(w_\text{avg}^{(t-1)})$
            \State Compute content loss $\mathcal{L}_\text{content}(x_0, x_\text{avg}^{(t-1)})$ using Eq.~\ref{eq:loss-content} 
            \State Compute local style alignment loss $\mathcal{L}_\text{style}((x'_1,\dots,x'_k), x_\text{avg}^{(t-1)})$ using Eq.~\ref{eq:loss-style}
            \State Compute total loss $\mathcal{L}_\text{total}=\lambda \mathcal{L}_\text{content}+(1-\lambda)\mathcal{L}_\text{style}$
            \State Update $w_{\text{avg}}^{(t)}$ using $w_{\text{avg}}^{(t-1)}$ and the gradient $\nabla_{w_\text{avg}^{(t-1)}} \mathcal{L}_\text{total}$
        \EndFor
        \State Add $G(w_{\text{avg}}^{(T)})$ to $\tilde{X}$
        \EndFor
    \State    \Return $\tilde{X}$

    \end{algorithmic}
\end{algorithm}

While prior  $k$-Same approaches developed for facial images have considered the Euclidean average of latent codes within each cluster as the representative, we found that this straightforward approach leads to significant loss of detail in synthetic retinal images, where clinically relevant patterns such as hemorrhages and exudates are often omitted (see Fig.~\ref{fig:examples}).

We make the following two observations toward addressing this key limitation.
First, unlike facial images where the salient structural features (e.g. eyes and nose) generally appear in consistent regions within the image, which facilitates the disentanglement of latent features, important patterns in retinal images can appear in different areas and thus are more easily diluted when averaging the features in the latent space. Second, in contrast to the importance of shape information in facial images, the patterns of interest in retinal images that are not directly linked to personal identity tend to be associated with \emph{texture-level} information (e.g. colored dots of varying granularity and frequency). In fact, the geometric structure of the blood vessels is a prominent identifying feature of concern that we are interested in obfuscating in the image via averaging.

Our local style alignment technique takes advantage of these observations to obtain higher quality representative images of each cluster.
We first draw the connection between texture patterns of interest and the ``style'' of the image from the style transfer literature~\cite{Gatys16}. Since we are interested in local texture patterns in the image, we capture the \emph{local style features} by constructing the feature covariance matrix in a sliding window of image patches, rather than over the whole image. 
We then consider the correspondence of the local style features
between the source images in the cluster and the target representative image, allowing for source texture patterns to appear in different locations in the target and simultaneously enforcing that these patterns are recapitulated somewhere in the target image.
Optimizing the latent code for the representative image with respect to the augmented loss function considering both the local style features and general content similarity, we obtain an enhanced representative image for each cluster. We provide the details of each step below and in Algorithm~\ref{alg:overall}.

\subsubsection{Construction of local style features.}

Following the approach of style transfer, we view style and texture information as being captured by the cross-channel feature correlations in the intermediate layers of a pre-trained convolutional neural network (CNN), such as VGG19~\cite{Simonyan15}. 

Formally, let $F_{\text{source}}^{(i)}$ and $F_{\text{target}}$ be $n$-by-$n$-by-$c$ tensors for the $i$-th source image and the target representative, respectively, representing the activation output of an $n$-by-$n$ intermediate CNN layer across $c$ channels. Note that we use the second layer of VGG19 in all our experiments. We spatially partition the activation output into a grid of $p$ submatrices, $\{F_{\text{source},j}^{(i)}\}_{j=1}^p$ and $\{F_{\text{target},j}\}_{j=1}^p$, each corresponding to a local patch in the image. For each patch $j$, we define the local style features $S^{(i)}_{\text{source},j}$ and $S_{\text{target},j}$ as $c$-by-$c$ matrices, where
\begin{align}
    (S^{(i)}_{\text{source},j})_{u,v} := \langle \textsf{Vec}((F^{(i)}_{\text{source},j})_{:,:,u}), 
    \textsf{Vec}((F^{(i)}_{\text{source},j})_{:,:,v})
    \rangle\\
    (S_{\text{target},j})_{u,v} := \langle \textsf{Vec}((F_{\text{target},j})_{:,:,u}), 
    \textsf{Vec}((F_{\text{target},j})_{:,:,v})
    \rangle
\end{align}
and $\textsf{Vec}(\cdot)$ denotes vectorization,  $\langle\cdot,\cdot\rangle$ denotes dot product, and $M_{:,:,u}$ for a tensor $M$ denotes a slice corresponding to channel $u$. 
The sets $\{S_{\text{source},j}^{(i)}\}_{j=1}^p$ for each source image $i$ in the cluster and $\{S_{\text{target},j}\}_{j=1}^p$ for the target fully characterize the local texture information we aim to capture in our model.

\subsubsection{Alignment of local style features.}
To introduce flexibility in determining where a visual pattern from the source image appears in the target image, we quantify the agreement in local style features via a correspondence. 
Inspired by the recent work of Wang et al. on dense contrastive learning~\cite{Wang21}, we compute the cosine similarity of style features between every pair of patches between the source and the target and take the optimal match for each patch in the source image to be included in the loss function. This induces positional flexibility while penalizing complete omission of texture patterns from the source. Note that the prior work~\cite{Wang21} did not consider style information in their approach.

We define correspondence index $a(i,j)$ for patch $j$ in the source image $i$ as:
\begin{equation}
    a(i,j) := {\arg\max}_{j'\in \{1,\dots,p\}} \textsf{CosineSimilarity}(\textsf{Vec}(S^{(i)}_{\text{source},j}), \textsf{Vec}(S_{\text{target},j'})).
\end{equation}
It is worth noting that, while a na\"{i}ve implementation of all pairwise comparison of patches leads to significant runtime overhead, our implementation efficiently utilizes matrix operations to maintain computational efficiency.

\subsubsection{Optimization of representative images.}
To synthesize an informative representative image for each cluster, leveraging the correspondence of local style features, we frame the process as an optimization problem as follows.

Let $E(\cdot)$ and $G(\cdot)$ be the encoder of the pre-trained GAN inversion model and the pre-trained GAN generator, respectively.
We directly optimize the target latent code $w_\text{avg}$ whose corresponding image $G(w_\text{avg})$ is the desired representative of the cluster.

We initialize $w_\text{avg}$ to the baseline Euclidean average $w_0$ of the source image embeddings given by 
\begin{equation}
    w_0 := \frac{1}{k} \sum_{i=1}^k E(x^{(i)}),
\end{equation}
where $x^{(i)}$ denotes the $i$-th image in the cluster.
We then iteratively optimize the solution using gradient descent with the loss function
\begin{equation}
\label{eq:loss}
    \mathcal{L}_\text{total} = \lambda \mathcal{L}_\text{content} + (1-\lambda) \mathcal{L}_\text{style},
\end{equation}
where $\lambda$ determines the ratio between the two terms given by 
\begin{align}
        \mathcal{L}_\text{content} &= 1-\langle F(G(w_0)), F(G(w_\text{avg})) \rangle,  \label{eq:loss-content} \\
        \mathcal{L}_\text{style} &= \sum_{i=1}^k \sum_{j=1}^p \| S^{(i)}_{\text{source},j} -  S_{\text{target},a(i,j)} (w_\text{avg}) \|_\mathcal{F}^2. \label{eq:loss-style}
\end{align} 
Note that $\|\cdot\|_\mathcal{F}$ denotes the Frobenius norm, and $F(\cdot)$ represents a pre-trained encoder network, which we use to induce high-level similarity between the optimized representative and the Euclidean average to avoid degenerate cases and to prioritize refining of the baseline solution.
In our experiments, we set $F$ to the pre-trained MoCo network~\cite{He20}, which was trained on the ImageNet dataset~\cite{Fei09} via unsupervised contrastive learning.
MoCo is recognized for its effectiveness in capturing semantic information of natural images beyond the available labels in the original dataset.
We also note that, although style transfer approaches typically take many iterations to converge, our initialization scheme using the Euclidean average greatly simplifies this process, requiring fewer iterations to obtain the final solutions in our experiments.

\subsection{GAN-based Image Generation and Encoding}
We train a GAN model, StyleGAN2-ADA~\cite{Karras20b}, on retinal images to learn to generate realistic fundus images from a latent vector space.
The StyleGAN family of methods~\cite{Karras20b,Karras19,Karras20a} generate high-resolution images using a progressive architecture, where increasingly fine-grain details are added to the image as we get deeper into the network.
We consider the latent space associated with this network to be extended multi-scale $\mathcal{W}$, which modulates the activation of units in all layers of the generator hierarchy.

StyleGAN2-ADA is one of the latest in this class, which stabilizes training on limited data using the \emph{adaptive discriminator augmentation} (ADA) mechanism.
Likely because the size of the public retinal image datasets is small compared to other types of image datasets, we observed that the use of ADA leads to a considerable improvement in image quality. 
We also note that the existing work on GANs for retinal images (e.g.~\cite{Chen21,Niu19,Yu19,Zhou20}) leverages additional labeled information such as vessel segmentation, and thus are not directly applicable to our setting where we use the raw fundus images.  

To manipulate and summarize real retinal images in the latent space equipped with a generator to synthesize new images, we need an encoder to map a given image to the GAN latent space, a task known as GAN inversion~\cite{Xia21}.

In our framework, we use this encoder to invert every image in the input dataset, then use the latent codes both to define the clusters of size $k$ to be averaged and to find the Euclidean centroid for the cluster to use as an initialization point, as described in the previous sections.

To this end, we use ReStyle~\cite{Alaluf21}, a recently developed approach to GAN inversion which achieved a significant scalability improvement over the previous methods by adopting an iterative refinement approach.
The ReStyle model takes the target image and the current synthetic image (the result of inversion) as input, and learns to generate an update to the latent code that improves consistency between the two images.
For $k$-SALSA, adopting this approach was key to building a practical pipeline---it reduced the inversion time by an order of magnitude (from 80 to 3 seconds per image).

While the generator and the encoder individually draws from prior work, we note that the combination of these state-of-the-art techniques have not been previously studied in the context of privatizing retinal images and were in fact key enabling factors of the practical performance of $k$-SALSA in our experiments.

\subsection{Same-Size Clustering}

To partition the input dataset into groups of exactly $k$ images to average, we employ a greedy nearest neighbor clustering to the inverted latent codes of the input images.
At each iteration, a point with the maximum average distance to the rest of the dataset is chosen, with the goal of prioritizing outliers.
Then $k-1$ nearest neighbors of the chosen point are identified to form a new cluster of size $k$.
The points in the new cluster are removed from the dataset and the above process is repeated.
In our experiments, we downsample the dataset to a multiple of $k$ to avoid a leftover cluster smaller than $k$.
Our experiments show the effectiveness of this efficient clustering approach in downstream tasks.

\section{Experiments}

\begin{figure*}
    \centering
    \includegraphics[width=.8\textwidth]{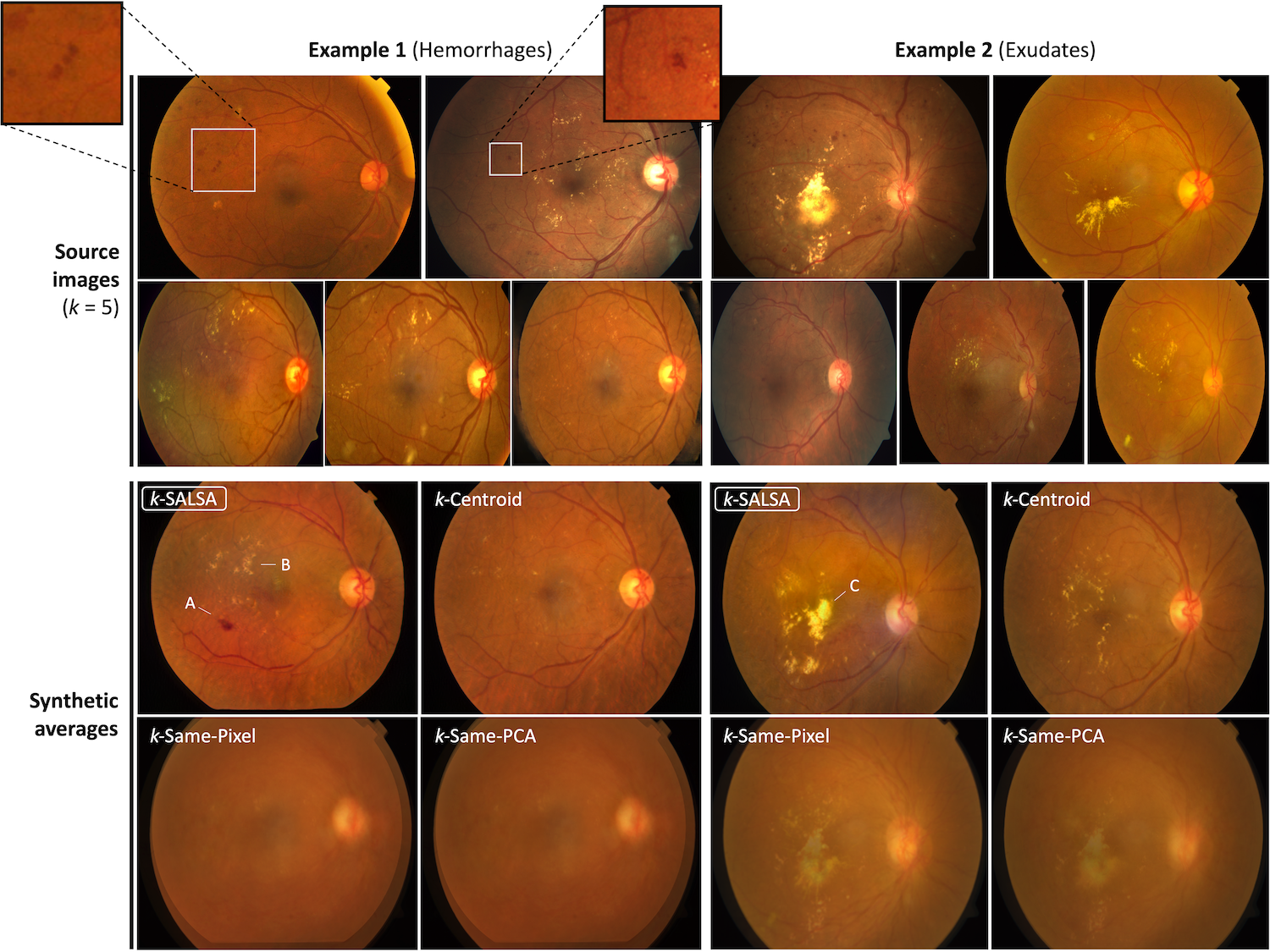}
    \caption{\textbf{Examples of synthetic average of retinal images.} 
    Two example clusters ($k=5$) of real retinal images (\emph{top}) along with synthetic averages generated by different methods (\emph{bottom}). $k$-SALSA better captures clinically relevant features such as hemorrhages (\emph{A}) and exudates (\emph{B} and \emph{C})}
    \label{fig:examples}
\end{figure*}

\subsection{Benchmark Datasets and Evaluation Setting}

Our experiments are conducted on public fundus image datasets APTOS\footnote{\url{https://www.kaggle.com/c/aptos2019-blindness-detection}} and EyePACS\footnote{\url{https://www.kaggle.com/c/diabetic-retinopathy-detection}}, widely used for diabetic retinopathy (DR) classification.
The images in both datasets are labeled by ophthalmologists with five grades of DR based on severity: 0 (normal), 1 (mild DR), 2 (moderate DR), 3 (severe DR), and 4 (proliferative DR).
EyePACS images were acquired from different imaging devices, leading to variations in image resolution, aspect ratio, intensity, and quality.
Hence, EyePACS represents a more challenging evaluation setting.

For both datasets, we train the GAN generator and the GAN inversion model on the training set (see supplement for details).
We then apply $k$-SALSA to the training set with the pre-trained GAN models to generate a $k$-anonymous dataset of synthetic images with aggregated labels.
To evaluate the downstream utility of the synthetic dataset, we trained DR classifiers based on the synthetic images and evaluated the classifiers on the test set using real images and labels.

\subsection{Baseline Approaches}
We compare $k$-SALSA with the following baseline methods. 
To demonstrate the advantage of our novel local style alignment-based averaging scheme, we consider the same method as $k$-SALSA, except using the Euclidean average (centroid) of each cluster in our GAN latent space to generate the representative image (\textbf{$k$-Centroid}).
To illustrate the importance of GANs in synthetic averaging of retinal images, we also evaluate less sophisticated schemes based on pixel-wise averaging  (\textbf{$k$-Same-Pixel}) and averaging in the low-dimensional latent space obtained by principal components analysis (\textbf{$k$-Same-PCA}).
Note that $k$-Same-PCA is equivalent to the method proposed in the original work on $k$-Same algorithms~\cite{Newton05}, and $k$-Centroid represents the best achievable performance following the general framework of $k$-Same-Net~\cite{Meden18} leveraging our GAN approaches.
We applied all averaging methods to the same set of clusters we constructed using the latent space of $k$-SALSA.
For some comparisons, we also consider the performance based on the non-averaged synthetic images generated from the inversion of each original image, i.e. $G(E(x))$ given an image $x$ (\textbf{GAN-Inverted}).

\subsection{Fidelity of Synthetic Images}
To evaluate the visual quality of synthetic retinal images, we first compare the Fr\'{e}chet inception distance (FID)~\cite{Heusel17}, a standard performance metric for images generated using GANs, across the methods we considered. Note that, unlike pixel-wise metrics such as PSNR and SSIM~\cite{wang2004image} (see supplement for additional discussion), FID measures the divergence between the multivariate Gaussian distributions induced by the real vs. synthetic images in the activation of the Inception V3 model~\cite{Szegedy16} trained on ImageNet~\cite{Fei09}. 
Intuitively, FID quantifies how different the synthetic images are from a reference set of real images in a manner that reflects human visual perception.
We use the original retinal images from each dataset as the reference to calculate FID on the corresponding synthetic images.
As shown in Table~\ref{table:fidelity}, $k$-SALSA consistently obtains the best (the lowest) FID among all averaging methods for different values of $k$ (2, 5, and 10) on both datasets.
For all methods, averaging leads to worse FID relative to GAN-Inverted images, with the gap increasing as $k$ becomes larger. This suggests that generating a realistic image becomes more difficult as we average more images.
Nevertheless, $k$-SALSA's FID remains closest to GAN-Inverted even for $k=10$.

\setlength{\tabcolsep}{4pt}

\begin{table}
\scriptsize
    \begin{center}
    \caption{Comparison of fidelity of synthetic images}
    \label{table:fidelity}
    \begin{tabular}{c|c|ccc|ccc}
        \toprule
        \multirow{2}{*}{\textbf{Method}} & \multirow{2}{*}{\textbf{{Metric}}} &\multicolumn{3}{c}{\textbf{APTOS}} & \multicolumn{3}{c}{\textbf{EyePACS}} \\
        & & $k=2$ & $k=5$ & $k=10$  & $k=2$ & $k=5$ & $k=10$  \\

        \midrule
        \midrule

        \multirow{1}{*}{GAN-Inverted} 
        & FID & {11.47} & {18.12} & {25.45} 
        & {9.65} & {14.65} & {19.77} \\
        \midrule
        
        \multirow{1}{*}{$k$-Same-Pixel} 
        & FID & {128.23} & {131.69} & {138.89} 
        & {65.21} & {113.45}  & {144.13} \\
        \midrule
        
        \multirow{1}{*}{$k$-Same-PCA} 
        & FID & {131.691} & {128.225} & {138.891} 
        & {159.57} & {164.69} & {173.15} \\
        \midrule

        \multirow{1}{*}{$k$-Centroid} 
        & FID & {12.7} & {22.71} & {31.09} 
        & {11.24} & {20.58} & {28.49} \\

        \midrule

         \multirow{1}{*}{\textbf{$k$-SALSA}} & FID & {\textbf{11.84}}& {\textbf{20.09}} & {\textbf{28.4}} 
         & {\textbf{9.95}} & {\textbf{15.07}} & {\textbf{21.28}} \\
        \bottomrule
    \end{tabular}
    \end{center}
\end{table}
\setlength{\tabcolsep}{1.4pt}

In Fig.~\ref{fig:examples}, we provide examples of synthetic averages generated by different methods for $k=5$ for visual comparison.
$k$-SALSA images more clearly capture the fine-grain, clinically-relevant patterns in the source images, including exudates (appearing as grainy yellow patches) and hemorrhages (appearing as dark spots), both of which are well-established biomarkers of diabetes~\cite{Mohamed07}.
$k$-Centroid generates realistic images, but tend to omit important fine-grain patterns, which initially motivated this work.
$k$-Same-Pixel and $k$-Same-PCA lead to low-fidelity images that even fail to align the boundaries of the photographs due to their linearity.
Examples for other values of $k$ are provided in the supplement.

\subsection{Downstream Classification Performance}

In addition to generating more realistic and informative summaries of each cluster of retinal images, we are interested in enabling downstream analysis with our synthetic data. We tested whether $k$-SALSA's synthetic dataset can lead to accurate classifiers of clinical labels, in our case the grading of diabetic retinopathy (DR). In EyePACS, we evaluated normal vs. DR binary classification due to the highly imbalanced number of labels (in contrast to the five-class setting in Kaggle).
We tested multi-class prediction with all five labels in APTOS.

The number of images in the training set was 3,000 for APTOS and 10,000 for EyePACS, where the latter was subsampled for efficiency. For each set, we used our clustering approach to obtain same-size clusters for each of $k\in \{2,5,10\}$, which were then individually averaged to obtain training images for a classifier.

We also evaluated the classifiers trained on original or GAN-Inverted images, which were subsampled to the same number of training examples as the synthetic datasets for each $k$ for comparison. We provide experimental details and a comparison without subsampling in the supplement.

\setlength{\tabcolsep}{4pt}
\begin{table}
\scriptsize
    \begin{center}
    \caption{Comparison of diabetic retinopathy classification performance}
    \label{table:classification}
    \begin{tabular}{c|c|ccc|ccc}
        \toprule
        \multirow{2}{*}{\textbf{Method}} & \multirow{2}{*}{\textbf{{Metric}}} &\multicolumn{3}{c}{\textbf{APTOS}} & \multicolumn{3}{c}{\textbf{EyePACS}} \\
        & & $k=2$ & $k=5$ & $k=10$  & $k=2$ & $k=5$ & $k=10$  \\
        \midrule
        \midrule
        \multirow{2}{*}{Original}  
        & Accuracy & {0.771} & {0.752} & {0.700} 
        & {0.794} & {0.767} & {0.725} \\
        & Cohen's $\kappa$ & {0.903} & {0.888} & {0.856} 
        & {0.414} & {0.327} & {0.297} \\
         \midrule
        
        \multirow{2}{*}{GAN-Inverted} 
        & Accuracy & {0.744} & {0.702} & {0.651} 
        & {0.731} & {0.730} & {0.708} \\
        & Cohen's $\kappa$ & {0.865} & {0.828} & {0.814} 
        & {0.140} & {0.140} & {0.058} \\
        \midrule
        
        \multirow{2}{*}{$k$-Same-Pixel} 
        & Accuracy & {0.502} & {0.318} & {0.366} 
        & {0.361} & {0.296}  & {0.434} \\
        & Cohen's $\kappa$ & {0.663} & {0.273} & {0.140} 
        & {0.029} & {0.000} & {0.043} \\
        \midrule
        
        \multirow{2}{*}{$k$-Same-PCA} 
        & Accuracy & {0.572} & {0.394} & {0.293} 
        & {0.3360} & {0.361} & {0.657} \\
        & Cohen's $\kappa$& {0.651} & {0.559} & {0.253} 
        & {0.010} & {0.029} & {0.096} \\
        \midrule
        
        \multirow{2}{*}{$k$-Centroid} 
        & Accuracy & {\textbf{0.688}} & {0.688} & {0.646} 
        & {0.680} & {0.664} & {0.611} \\
        & Cohen's $\kappa$ & {\textbf{0.786}} & {0.745} & {0.647} 
        & {\textbf{0.268}} & {0.185} & {0.160}  \\
        
        \midrule

         \multirow{2}{*}{\textbf{$k$-SALSA}} & Accuracy & {0.687}& {\textbf{0.712}} & {\textbf{0.673}} 
         & {\textbf{0.704}} & {\textbf{0.688}} & {\textbf{0.705}} \\
        & Cohen's $\kappa$ & {0.773} & {\textbf{0.769}} & {\textbf{0.710}} 
        & {0.254} & {\textbf{0.222}} & {\textbf{0.225}}   \\
        \bottomrule
    \end{tabular}
    \end{center}
\end{table}
\setlength{\tabcolsep}{1.4pt}

The results in Table~\ref{table:classification} show that the $k$-SALSA synthetic datasets generally outperform the alternative approaches with respect to both accuracy and Cohen's $\kappa$ statistic (with quadratic weighting) on the test set.
$k$-Centroid achieves slightly better performance for $k=2$, but remains comparable to our approach.
Since Euclidean averaging may introduce greater distortions for larger values of $k$, we expect the advantage of $k$-SALSA to be more pronounced for moderate to large $k$, which is consistent with our results.
As expected, performance based on original images is higher than the synthetic dataset, but part of this gap is ascribed to the limitations of the current GAN models as suggested by the lower performance of the non-averaged, GAN-inverted images compared to the original.
Interestingly, for EyePACS $k = 10$, $k$-SALSA outperforms GAN-Inverted, suggesting that summarizing salient features may even be beneficial for classification when the data is limited.

We include in the supplement additional results illustrating the impact of $k$ and a promising extension of $k$-SALSA which uses data augmentation to mitigate dataset reduction due to averaging.

\subsection{Mitigation of Membership Inference Attacks}

To compare the privacy properties of the methods, we implemented a membership inference attack (MIA), where an adversary holding a synthetic dataset attempts to infer whether a target person was part of a specific cluster. 
We trained ResNet18~\cite{He16} on the training set to classify cluster membership using the synthetic averages generated by each method. We then evenly divided the test set into two parts, generated cluster averages on the first, then evaluated the performance of the classifier in ranking the images in \emph{both} test sets for membership in each cluster, based only on its synthetic average.
For each cluster size $k$, we calculated the top-$K$ accuracy (i.e., mean fraction of top $K$ samples in the ranking that correspond to correct guesses) with $K=k$.

\setlength{\tabcolsep}{4pt}
\begin{table}
\scriptsize
    \begin{center}
    \caption{Membership inference attack top-$K$ accuracy (\%) with $K=k$.}
    \label{table:privacy}
    \begin{tabular}{c|ccc|ccc}
        \toprule
        \multirow{2}{*}{\textbf{Method}}  &\multicolumn{3}{c}{\textbf{APTOS}} & \multicolumn{3}{c}{\textbf{EyePACS}} \\
        & $k=2$ & $k=5$ & $k=10$  & $k=2$ & $k=5$ & $k=10$  \\

        \midrule
        \midrule
        
        \multirow{1}{*}{$k$-Same-Pixel} 
         & {100.0} & {91.76} & {84.21} 
        & {97.96} & {94.28} & {86.67} \\
        \midrule
        
        \multirow{1}{*}{$k$-Same-PCA} 
         & {98.98} & {84.29} & {2.5} 
        & {98.98} & {86.21} & {2.63} \\
        \midrule
        
        \multirow{1}{*}{$k$-Centroid} 
         & {78.57} & {41.42} & {2.1} 
        & {88.63} & {45.92} & {2.63} \\
        
        \midrule
         \multirow{1}{*}{\textbf{$k$-SALSA}} 
         & {\textbf{77.55}} & {\textbf{40.0}} & {\textbf{1.0}}
        & {\textbf{71.42}} & \textbf{{35.17}} & \textbf{{0.52}} \\
        \bottomrule
    \end{tabular}
    \end{center}
\end{table}
\setlength{\tabcolsep}{1.4pt}

The results are summarized in Table~\ref{table:privacy}.
Note that an adversary with access to the encoder would achieve an expected accuracy of 50\% for all values of $k$, since in any neighborhood a random half of the samples correspond to negative matches that were not included in the private dataset. 
This represents a realistic scenario where the attacker does not have \emph{a priori} knowledge of individuals in the private dataset.
For a worst-case evaluation, we assume that the target's image is identical to the one in the dataset; any protection offered by noisy re-acquisition of images is likely to be bypassed with more sophisticated MIA (e.g., using vessel structures).
We observed that pixel-averaging provides little to no privacy.
$k$-Same-PCA and $k$-Centroid lower the risks, the latter to a greater extent.
$k$-SALSA results in the strongest mitigation with MIA accuracy of 1\% and 0.52\% for APTOS and EyePACS, respectively, for $k=10$.
Our improvement over $k$-Centroid is likely due to the fact that the addition of our local style loss prioritizes similarity in high-level visual patterns over low-level content, potentially reducing the amount of identity-related information that can be exploited by the attack.
None of the methods provide strong privacy at $k=2$, which reflects an insufficient amount of variability between the two source images that could be leveraged for privacy; however, we expect our approach to provide meaningful privacy protection for larger values of $k$ as our results show.

\subsection{Ablation Study}
We conducted an ablation study to evaluate the importance of individual components of our methodology.
Recall that the loss function of $k$-SALSA includes the content loss and the local style loss (see Eq.~\ref{eq:loss}).
We considered four alternative models with: only style loss, only content loss, both but using global style features computed over the whole image, and both without the flexible alignment (i.e., each local style is directly compared to that of the corresponding patch in the other image at the same location).
All of these alternatives performed considerably worse than $k$-SALSA in downstream classification performance (Table~\ref{table:ablation}).
The especially poor performance without alignment suggests that enforcing style preservation without spatial flexibility can in fact be harmful for the method.
\begin{table}
\setlength{\tabcolsep}{4pt}
\scriptsize
    \begin{center}
    \caption{Ablation study (APTOS, $k=5$)}
    \label{table:ablation}
    \begin{tabular}{l|cc}
        \toprule
        \multirow{1}{*}{\textbf{Method}} & \multirow{1}{*}{\textbf{{Accuracy}}} &\multirow{1}{*}{\textbf{{Cohen's} $\kappa$}} \\

        \midrule
        \midrule

        \multirow{1}{*}{Style loss only (local, with alignment)} 
        & {0.685} & {0.735}  \\

        \midrule
        \multirow{1}{*}{Content loss only} 
        & {0.673}& {0.712}\\
        
        \midrule
        \multirow{1}{*}{Content loss, Global style loss}
        & {0.687} & {0.761}  \\

        \midrule
        \multirow{1}{*}{Content loss, Local style loss, No alignment}
        & {0.57} & {0.417}  \\

        \midrule
        \midrule

         \multirow{1}{*}{\textbf{$k$-SALSA}} 
        & {\textbf{0.712}} & {\textbf{0.769}}  \\
        \bottomrule
    \end{tabular}
    \end{center}
\end{table}

\section{Discussion and Conclusions}
We presented $k$-SALSA, an end-to-end pipeline for synthesizing a $k$-anonymous retinal image dataset given a private input dataset.
We leverage local style alignment, our new approach for summarizing source images in a cluster while preserving local texture information.
Our results demonstrate that $k$-anonymization of retinal images, preserving both privacy and clinical utility, is feasible.

We would like to address several limitations of the current method in future work. First, $k$-SALSA's performance is dependent on the quality of the underlying GAN and GAN inversion models.
We plan to devise strategies tailored to retinal images (e.g., separately modelling different parts of the image) to further improve GAN models. Next, we plan to explore more rigorous frameworks for privacy such as differential privacy (DP)~\cite{Dwork14}. While it is generally difficult to apply DP to high-dimensional data such as images, certain relaxations of DP~\cite{Kifer14} may lead to a practical solution. Lastly, we plan to explore the application of our methodology to other imaging modalities for the retina, including the OCT.

Our work demonstrates that domain-inspired techniques can be combined with the state-of-the-art GAN techniques to design effective approaches to privatizing sensitive data. 
The methodological insights introduced by our work is of general interest to other domains (e.g. genomics), where privacy-aware aggregation of sensitive data may overcome challenges in data sharing.\\

\noindent\textbf{Acknowledgements.} M.J. is supported by the Ministry of Trade, Industry, and Energy in Korea, under Human Resource Development Program for Industrial Innovation (Global) (P0017311) supervised by the Korea Institute for Advancement of Technology. H.C. is supported by NIH DP5 OD029574-01 and by the Schmidt Fellows Program at Broad Institute.

%
%
\bibliographystyle{splncs04}

\setcounter{section}{0}
\renewcommand\thesection{\Alph{section}}

\chapter*{Supplementary Materials}

\section{Benchmark datasets}
The APTOS dataset includes 3,662 labeled fundus images. 
We split the data into a training set of 3,000 images and a test set of 662 images.
The EyePACS dataset includes 35,126 labeled fundus images.
We split the data into a training set of 28,100 images and a test set of 7,026 images.
We rescaled the images in both datasets to 512-by-512 RGB pixels.
The training set is used to obtain the GAN and GAN-Inversion models.

\section{Implementation details}

\subsubsection{GAN.}
We trained our GAN models using the official PyTorch implementation\footnote{\url{https://github.com/NVlabs/stylegan2-ada-pytorch}} of StyleGAN2-ADA~\cite{Karras2020ada}. We set the number of mapping networks to two as recommended based on our image resolution and GPU count. Since the desired size of the generated images is $512 \times 512$, we used 16 progressive layers, resulting in the latent space $W$ with dimensions $16 \times 512$. We train the models on 5,000 kimgs in each dataset with batch size 64, using 8 3090-RTX GPUs, Pytorch 1.7.1, CUDA 11.1, and CuDNN 8.1.1.

\subsubsection{GAN Inversion.}
To obtain the GAN inversion encoder we built upon the official PyTorch implementation\footnote{https://github.com/yuval-alaluf/restyle-encoder} of ReStyle~\cite{Alaluf21}. We incorporated the MOCO-based~\cite{He20} similarity loss on pSp~\cite{richardson2021encoding} architecture with ResNetBackboneEncoder~\cite{he2016deep}. We trained each model for 100,000 iterations with a batch size of 8 and 5 refinement iterations per batch. Similar to the GAN setting, the output image size is set to $512\times 512$. We performed the training using 1 3090-RTX GPU with the same environment as that of GAN training.

\subsubsection{$k$-SALSA.}
We split the intermediate-level features into a grid of $4 \times 4$ patches (16 in total) to construct the local style features in all settings.
For the relative ratio of $\mathcal{L}_{\text{content}}$ and $\mathcal{L}_{\text{style}}$, we set the parameter $\lambda$ to 0.1, 0.05 and 0.03 for $k=2,5,10$, respectively, in APTOS. For EyePACS, we set $\lambda$ to 0.01, 0.02 and 0.01, respectively for each $k$.
We optimize our model for synthetic averaging using standard stochastic gradient descent with Adam~\cite{kingma2014adam}, with learning rate 0.1 and $\beta_1 = 0.9$, $\beta_2 = 0.99$. We use the same computational environment as above with a single GPU.

\subsubsection{Downstream classification.}
For all synthetic datasets, we trained a DR classifier using the ResNet50 model~\cite{He16} with batch size 32, 60 epochs, stochastic gradient descent (SGD) with Nesterov momentum 0.9~\cite{Nesterov03}, weight decay 0.0005, and cosine annealing in the learning rate schedule~\cite{Loshchilov16}.

\section{Computational costs}
One-time pre-training of GAN and inversion models took 10~hrs and 3~days, respectively, for APTOS, and  30~hrs and 3~days for EyePACS. 
The main runtime of $k$-SALSA depends on the inference speed of GAN and inversion, only around 0.85~secs/image.
Synthetic averaging takes 19~secs/cluster ($k$=5).
Both steps can be parallelized.
Same-size clustering takes $<$1~min.
Cosine similarity is computed for 16 patches/image (4x4) for 2~ms/image.
Overall, we expect $k$-SALSA to be practical in realistic settings. 

\section{Choice of similarity metric for local style alignment}
Here we provide additional empirical results supporting our choices of similarity metric in the local style alignment.
Recall that once the local style features are constructed for each batch, we find the optimal matches between the target and source images using cosine similarity (COS).
Given these optimal matches, we then use mean squared error (MSE) to optimize the synthetic average (see the definition of $\mathcal{L}_{\text{style}}$ in Eq.~7), effectively transferring the local styles from the source image to the target average.
We note that these choices are inspired by prior works; dense contrastive learning~\cite{Wang21} uses the COS metric to perform the alignment, whereas style transfer~\cite{Gatys16} is typically done using the MSE---our work combines these two approaches and keeps the respective similarity metrics.
As shown in Supplementary Table~\ref{table:ablation2}, using either of COS and MSE metrics for both components of the model (COS-COS or MSE-MSE) results in worse averaging performance as measured by the downstream classification task evaluated in our work, which intuitively captures how well the clinically relevant features are preserved in the averaged images.
This result supports our hybrid use of both metrics.

\setlength{\tabcolsep}{4pt}

\renewcommand{\tablename}{Supplementary Table}

\begin{table}
    \begin{center}
    \caption{Comparison with alternative similarity metrics in local style alignment with respect to classification performance (APTOS, $k=5$)}
    \label{table:ablation2}
    \begin{tabular}{l|cc}
        \toprule
        \multirow{1}{*}{\textbf{Method}} & \multirow{1}{*}{\textbf{{Accuracy}}} &\multirow{1}{*}{\textbf{{Cohen's} $\kappa$}} \\

        \midrule
        \midrule

        \multirow{1}{*}{MSE-MSE} 
        & {0.599} & {0.717}  \\

        \midrule
        \multirow{1}{*}{COS-COS} 
        & {0.708}& {0.727}\\

        \midrule
        \midrule

         \multirow{1}{*}{\textbf{$k$-SALSA} (COS-MSE)} 
        & {\textbf{0.712}} & {\textbf{0.769}}  \\
        \bottomrule
    \end{tabular}
    \end{center}
\end{table}
\setlength{\tabcolsep}{1.2pt}

\section{Addressing the reduced size of synthetic dataset}
Dataset size is an important issue in medical imaging problems.
The reduced number of images in the synthetic average dataset constructed by $k$-SALSA is a potential concern.
To mitigate the cost of dataset reduction, we investigated an extension of $k$-SALSA based on data augmentation, whereby small random noise is added to $k$-SALSA's average embeddings to generate multiple ``views'' of each cluster.
As shown in Supplementary Table~\ref{table:reduced}, with 5 augmented images per cluster, we observed an improved performance of 0.829 (originally 0.769) for $k$-SALSA, and 0.809 (0.745) for $k$-Centroid in APTOS, $k$=5. 
This demonstrates the potential of our extension in countering the size reduction of the synthetic dataset with data augmentation.
Importantly, the augmented images are independent of private images conditioned on $k$-SALSA's representative embedding of each cluster, thus there is no additional privacy leakage.
\setlength{\tabcolsep}{4pt}
\begin{table}
    \begin{center}
    \caption{Addressing the reduced size of synthetic dataset}
    \label{table:reduced}
    \begin{tabular}{c|c|ccc}
        \toprule
        \multirow{2}{*}{\textbf{Method}} & \multirow{2}{*}{\textbf{{Metric}}} &\multicolumn{2}{c}{\textbf{APTOS}, $k=5$} \\
        & & Augmented (5$\times$) & Non-Augmented \\
        \midrule
        {Centroid} & Cohen's $\kappa$ & {0.809} & {0.745} \\
        \textbf{$k$-SALSA} & Cohen's $\kappa$ & {0.829} & {0.769}\\
        
        \bottomrule
    \end{tabular}
    \end{center}
\end{table}
\setlength{\tabcolsep}{1.4pt}

\section{Additional comparisons with Original/gan-inverted}
In our main experiments, we subsampled the original and GAN-inverted images to match the number of training images for the classifiers for comparison with different synthetic average datasets.
For completeness, here we include the ``best-case scenario'' classification performance results for these baselines by using the full training dataset in APTOS. 
\setlength{\tabcolsep}{4pt}
\begin{table}
    \begin{center}
    \caption{Classification performance with and without subsampling and using data augmentation with $k$-SALSA}
    \label{table:fairness}
    \begin{tabular}{c|c|ccc}
        \toprule
        \multirow{2}{*}{\textbf{Method}} & \multirow{2}{*}{\textbf{{Metric}}} &\multicolumn{3}{c}{\textbf{APTOS}} \\
        & & Full & Subsampled ($k=5$) & Augmented  \\
        \midrule
        {Original} & Cohen's $\kappa$ & {0.914} & {0.888} & -\\
        {GAN-Inverted} & Cohen's $\kappa$ & {0.857} & {0.828} & - \\
        {$k$-SALSA} & Cohen's $\kappa$ & - & {0.769} & {0.829} \\
        
        \bottomrule
    \end{tabular}
    \end{center}
\end{table}
\setlength{\tabcolsep}{1.4pt}

The results are shown in Supplementary Table~\ref{table:fairness}.
We obtained a Cohen's $\kappa$ of 0.914 and 0.857 for Original and GAN-Inverted, respectively, without subsampling, compared to 0.888 and 0.828 with subsampling with $k=5$ (i.e., a 20\% sampling rate), respectively.
These results suggest that, while subsampling does reduce the performance, the impact is relatively small and that $k$-SALSA performance is still competitive with the best-case scenario, especially when used with our data augmentation strategy described in the previous section.

\section{Performance dependence on the cluster size $k$}
To further investigate the effect of cluster size $k$ on downstream classification performance, we compared the classifiers trained on $k$-SALSA synthetic datasets for different values of $k$, but subsampled to the same number of clusters.
Note that larger $k$ leads to fewer clusters and thus smaller training data for classification.
At the same time, larger $k$ increases the potential to retain more salient features from source images.
As shown in Supplementary Table~\ref{table:dependence_k}, we observed a performance improvement for larger $k$ in APTOS,
suggesting that summarizing key features across multiple images has a beneficial impact on the classifier training.
\setlength{\tabcolsep}{4pt}
\begin{table}
    \begin{center}
    \caption{Performance dependence on the cluster size $k$}
    \label{table:dependence_k}
    \begin{tabular}{c|c|ccc}
        \toprule
        \multirow{2}{*}{\textbf{Method}} & \multirow{2}{*}{\textbf{{Metric}}} &\multicolumn{3}{c}{\textbf{APTOS}} \\
        & & {$k=2$} & {$k=5$} & {$k=10$}  \\
        \midrule
        {$k$-SALSA} & Cohen's $\kappa$ & 0.688 & 0.740 & 0.761 \\
        
        \bottomrule
    \end{tabular}
    \end{center}
\end{table}
\setlength{\tabcolsep}{1.4pt}

\section{Challenges of differentially private GANs}
To evaluate the performance of differentially private GAN approach to synthetic generation of retinal images, we trained one of the state-of-the-art models, G-PATE~\cite{long2021gpate}, on our fundus image datasets (APTOS) using the official TensorFlow implementation\footnote{https://github.com/AI-secure/G-PATE}.
We follow the setting in the provided code except for the number of teacher networks and batch size, which we changed to 600 and 32 (from 2000 and 64), respectively, to reflect the smaller sizes of our datasets.
In order to use the provided code, we downscaled the retinal images to $64 \times 64$.
Otherwise, we used the following default parameters: number of epochs 1000, $\sigma$ threshold 600, $\sigma$ 100, step size $10^{-4}$, max $\epsilon$ 100 and $z$-dimension 100.
As shown in Supplementary Figure~\ref{fig:g_pate}, the generated images from the differentially private GAN, even with a lenient privacy parameter of $\epsilon=100$, are far from resembling retinal images.
We attribute this failure in training to the relatively small size of our datasets (e.g. 3000) and the high resolution of the images, compared to handwritten digit images considered in the original work.
Note that the GAN architecture used by G-PATE is DC-GAN~\cite{radford2015unsupervised}, which is expected to have difficulties in the limited-data, the high-resolution setting given its low representation power, instability, and vanishing gradients compared with more recent techniques such as StyleGAN2-ADA~\cite{Karras2020ada}.
Consistent with the visual assessment, the Fr\'{e}chet Inception Distance (FID) of the images generated by G-PATE is 441.23, which is vastly higher than that of our approach (20.09), indicating the challenges of differentially private training of GANs in our setting.

\renewcommand{\figurename}{Supplementary Figure\,}
\begin{figure*}
    \centering
    \includegraphics[width=.3\textwidth]{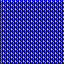}
    \caption{64 $\times$ 64 generated images from the G-PATE model~\cite{long2021gpate} trained on retinal images (APTOS).}
    \label{fig:g_pate}
    \vspace{-1.7em}
\end{figure*}

\section{Choice of visual fidelity metric}
To evaluate the fidelity of synthetic images, we use the Fr\'{e}chet inception distance (FID), a standard metric for images generated using GANs. Other common metrics include PSNR and SSIM; however, these metrics quantify the degradation of quality when a source image is transformed, whereas FID measures a distributional similarity to a set of reference images based on high-level activations. FID uniquely assesses whether $k$-SALSA images are realistic compared to real retinal images. Moreover, $k$-SALSA introduces spatial flexibility of image features, which is not captured by pixel-level metrics like PSNR and SSIM.

\section{Additional examples of synthetic averages}
\begin{figure*}
    \centering
    \includegraphics[width=1.0\textwidth]{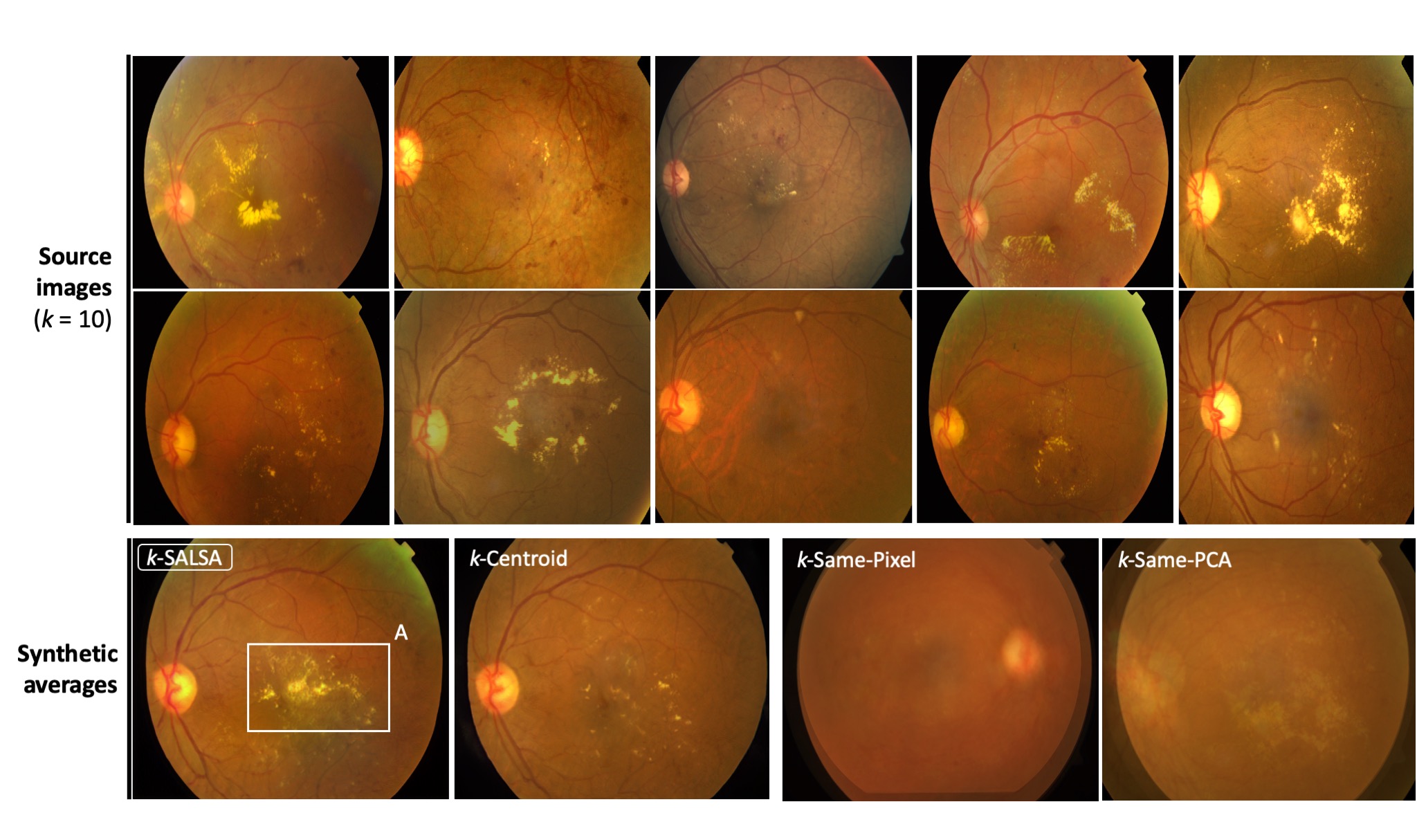}
    \caption{\textbf{Examples of synthetic average of retinal images ($k=10$). }
    One example of $k=10$ real images (\emph{top}) along with synthetic averages generated by different methods (\emph{bottom}). $k$-SALSA better captures a disease-related feature (\emph{A}).}
    \label{fig:k10_1}
    \vspace{-1.7em}
\end{figure*}

\begin{figure*}
    \centering
    \includegraphics[width=1.0\textwidth]{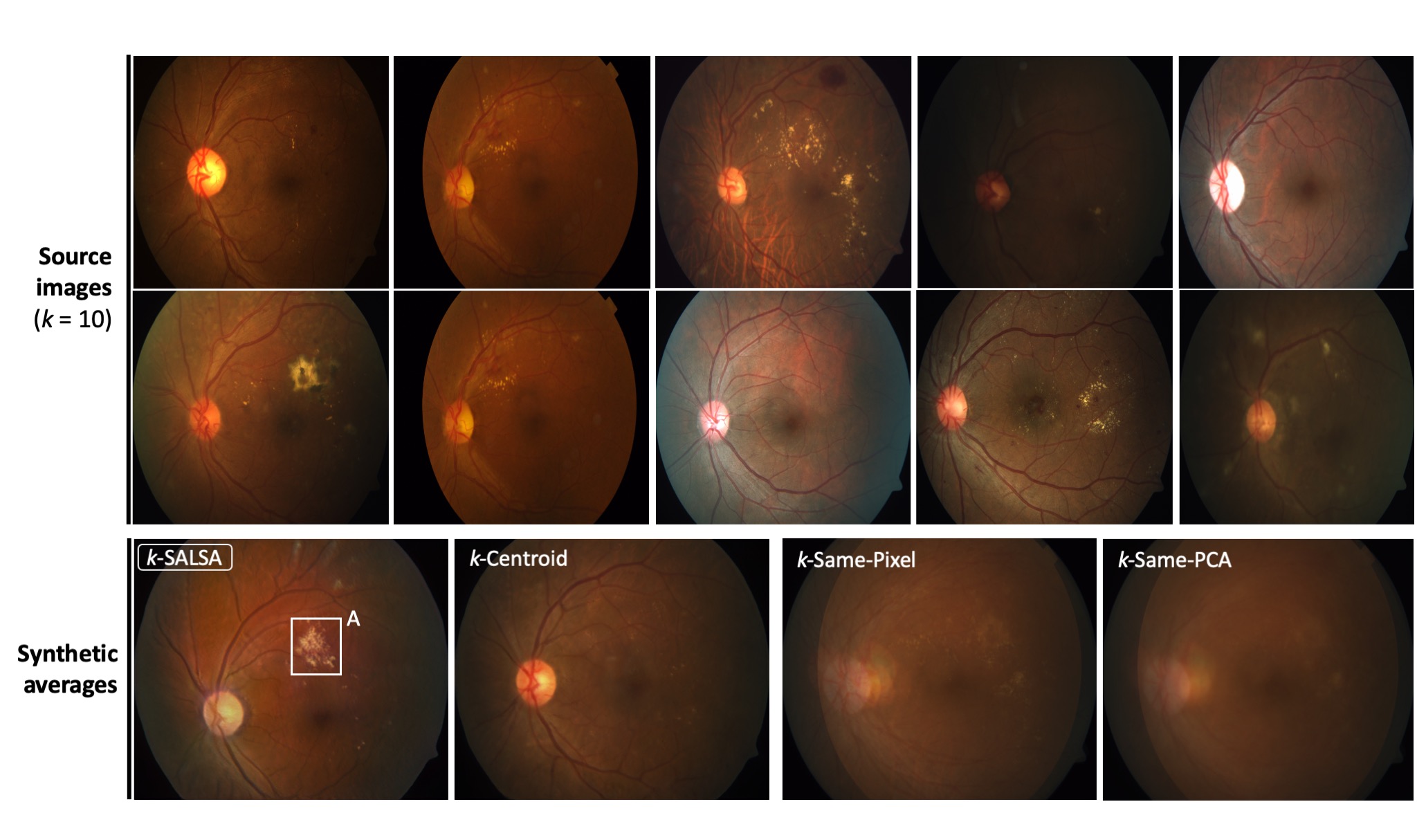}
    \caption{\textbf{Examples of synthetic average of retinal images ($k=10$).}
    One example of $k=10$ real images (\emph{top}) along with synthetic averages generated by different methods (\emph{bottom}). $k$-SALSA better captures a disease-related feature (\emph{A}).}
    \label{fig:k10_2}
    \vspace{-1.7em}
\end{figure*}

\begin{figure*}
    \centering
    \includegraphics[width=1.0\textwidth]{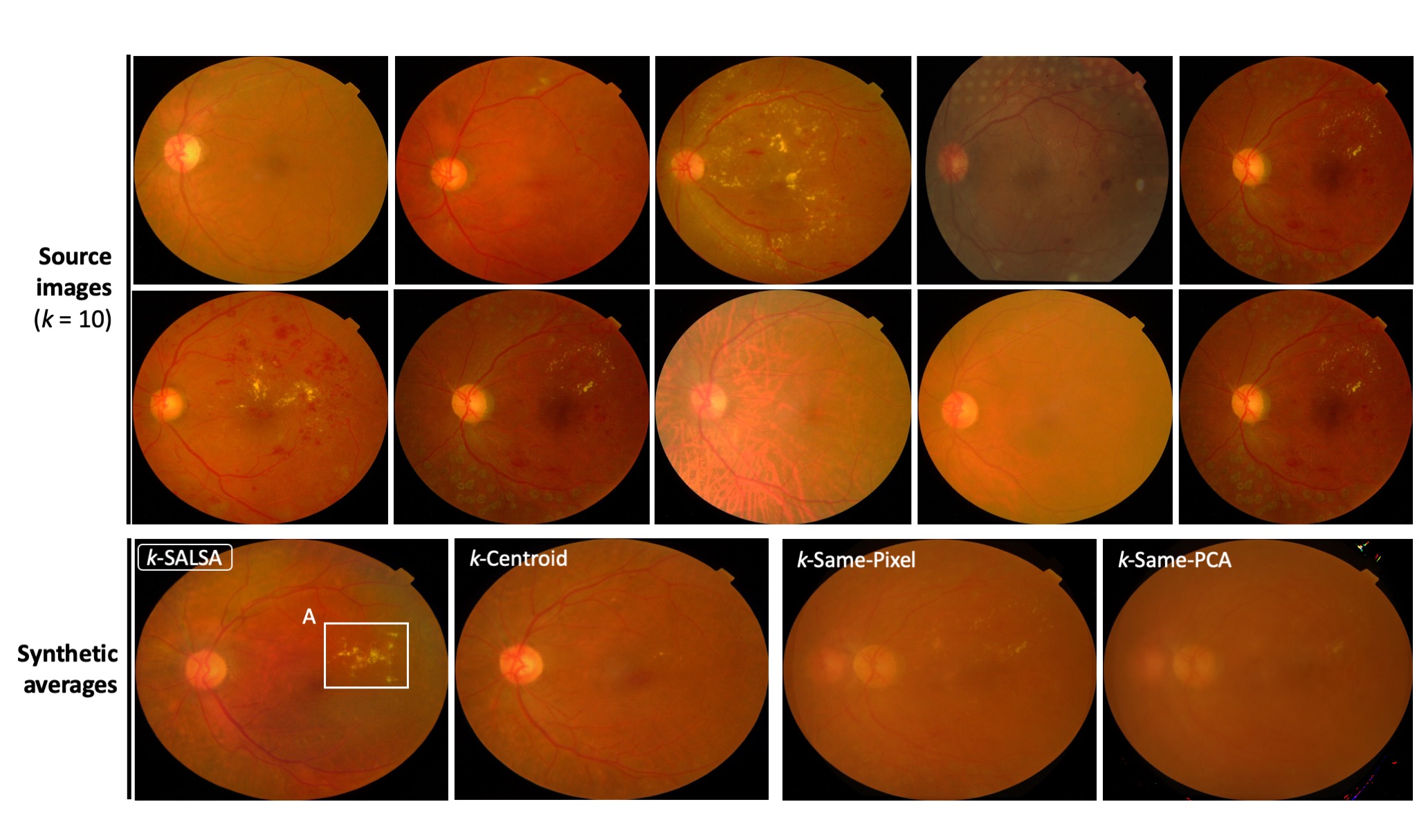}
    \caption{\textbf{Examples of synthetic average of retinal images ($k=10$).}
    One example of $k=10$ real images (\emph{top}) along with synthetic averages generated by different methods (\emph{bottom}). $k$-SALSA better captures a disease-related feature (\emph{A}).}
    \label{fig:k10_3}
    \vspace{-1.7em}
\end{figure*}

\begin{figure*}
    \centering
    \includegraphics[width=1.0\textwidth]{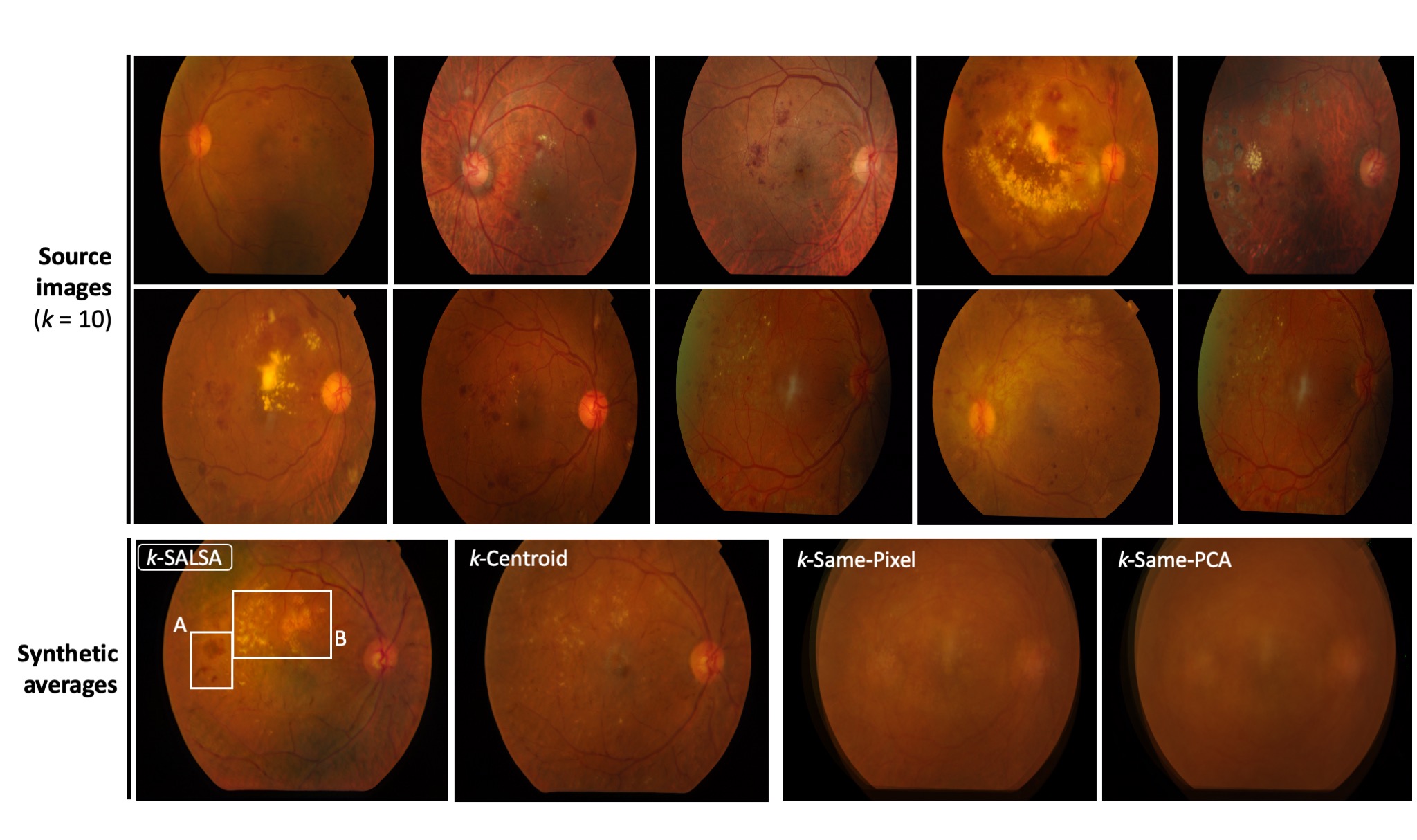}
    \caption{\textbf{Examples of synthetic average of retinal images ($k=10$).}
    One example of $k=10$ real images (\emph{top}) along with synthetic averages generated by different methods (\emph{bottom}). $k$-SALSA better captures  disease-related features (\emph{A}, \emph{B}).}
    \label{fig:k10_4}
    \vspace{-1.7em}
\end{figure*}

\begin{figure*}
    \centering
    \includegraphics[width=1.0\textwidth]{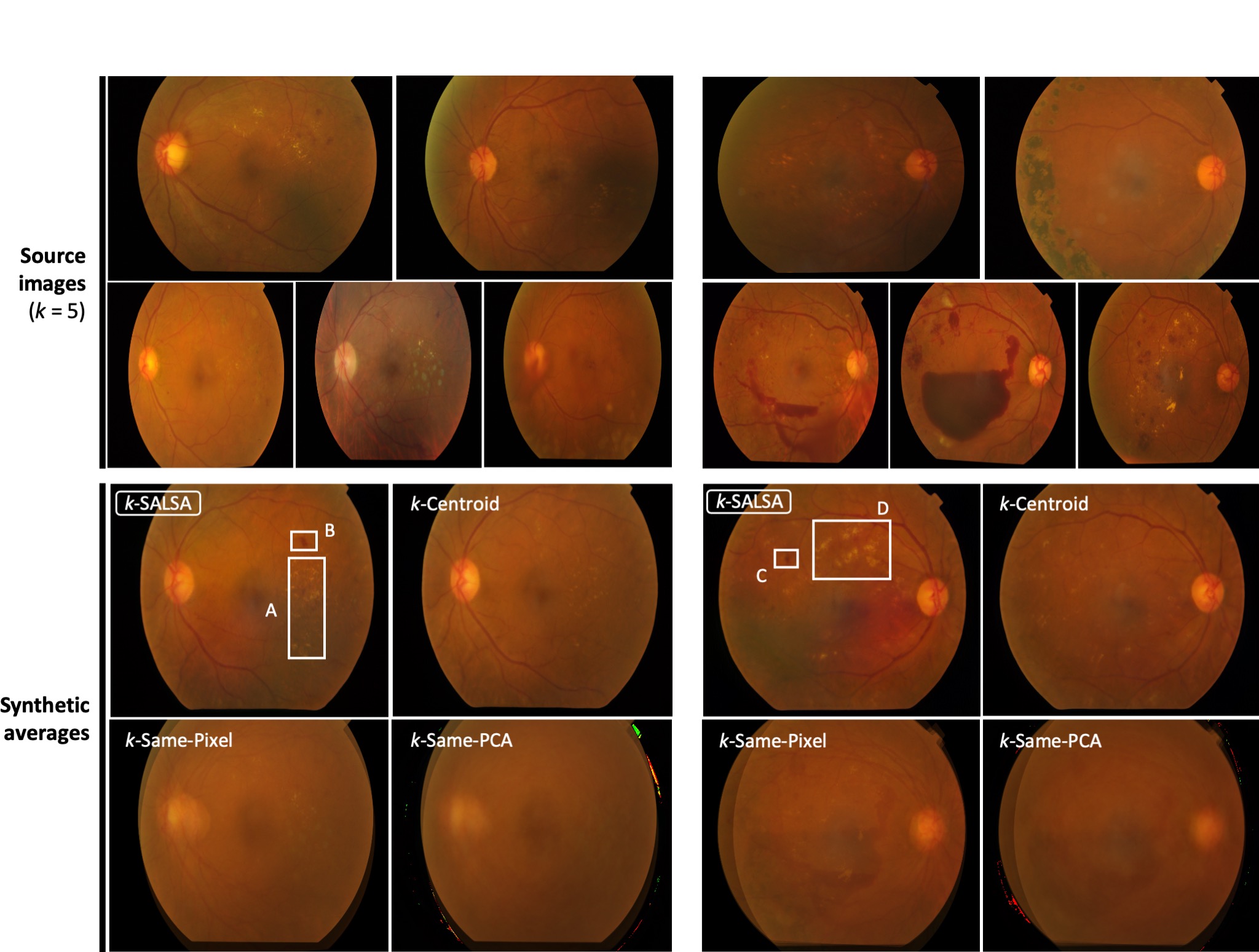}
    \caption{\textbf{Examples of synthetic average of retinal images ($k=5$).}
    Two examples of $k=5$ real images (\emph{top}) along with synthetic averages generated by different methods (\emph{bottom}). $k$-SALSA better captures  disease-related features (\emph{A}, \emph{B}, \emph{C}, \emph{D}).}
    \label{fig:k5_1}
    \vspace{-1.7em}
\end{figure*}

\begin{figure*}
    \centering
    \includegraphics[width=1.0\textwidth]{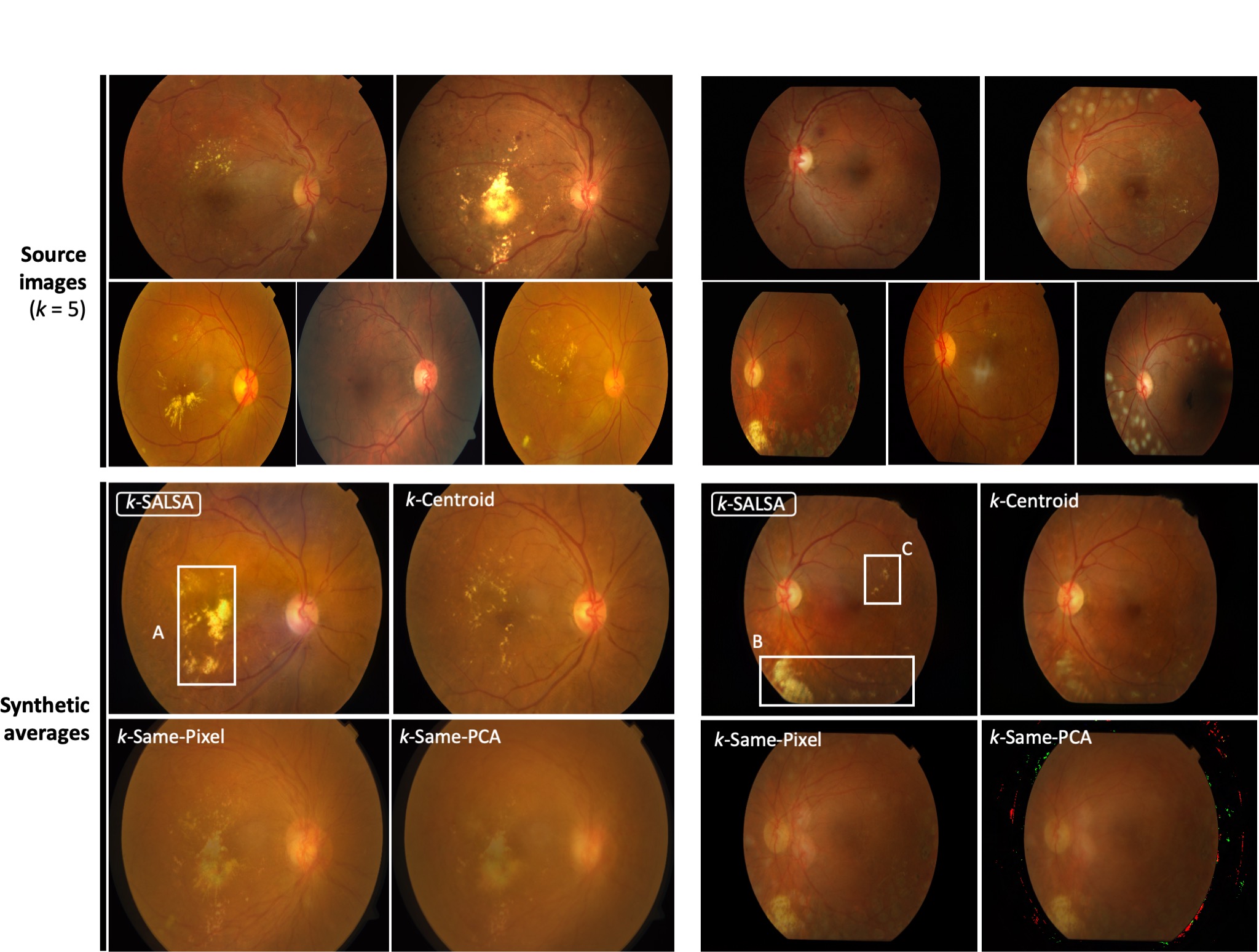}
    \caption{\textbf{Examples of synthetic average of retinal images ($k=5$).}
    Two examples of $k=5$ real images (\emph{top}) along with synthetic averages generated by different methods (\emph{bottom}). $k$-SALSA better captures disease-related features (\emph{A}, \emph{B}, \emph{C}).}
    \label{fig:k5_2}
    \vspace{-1.7em}
\end{figure*}

\begin{figure*}
    \centering
    \includegraphics[width=1.0\textwidth]{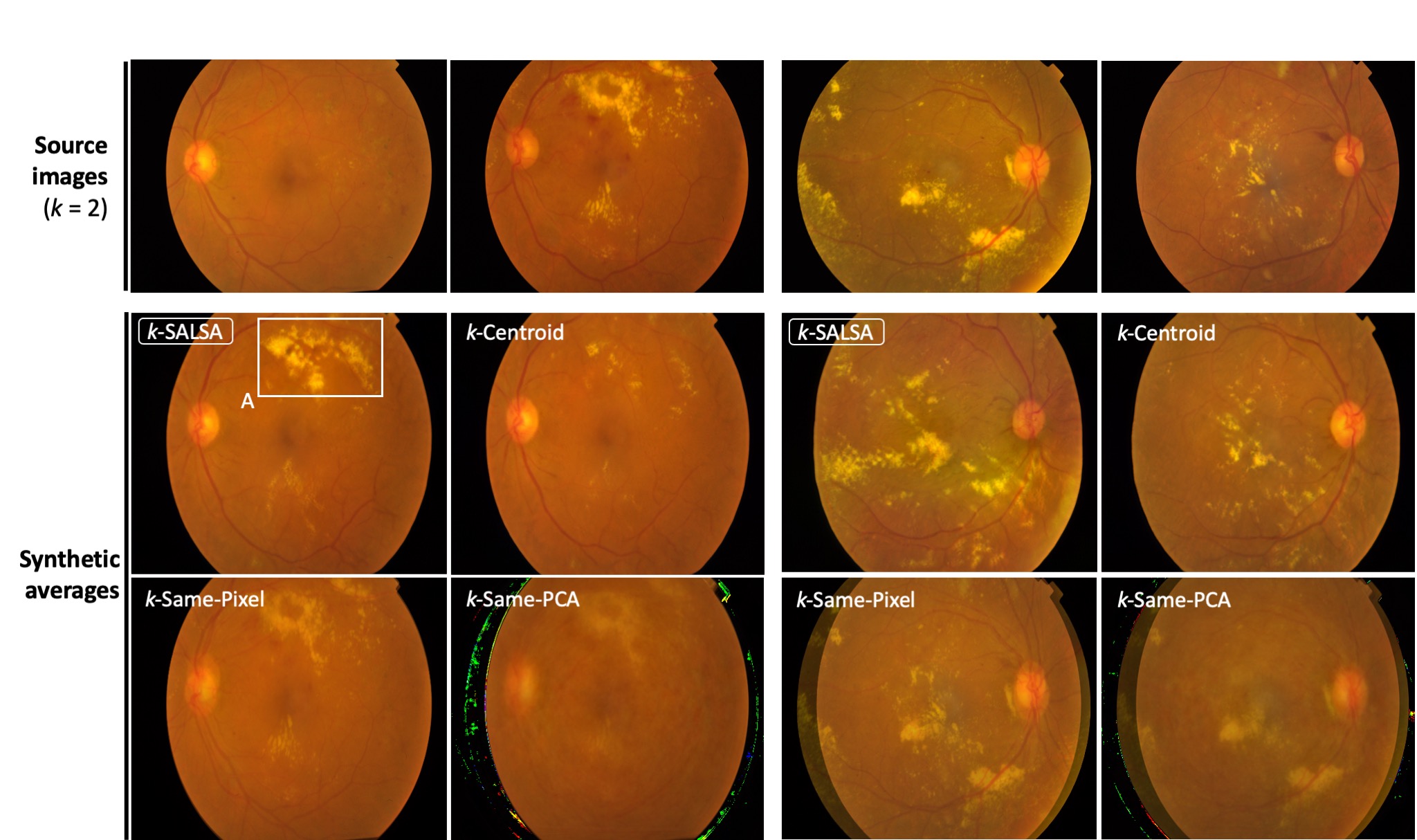}
    \caption{\textbf{Examples of synthetic average of retinal images ($k=2$).}
Two examples of $k=2$ real images (\emph{top}) along with synthetic averages generated by different methods (\emph{bottom}). $k$-SALSA better captures a disease-related feature (\emph{A}).}
    \label{fig:k2_1}
    \vspace{-1.7em}
\end{figure*}

\begin{figure*}
    \centering
    \includegraphics[width=1.0\textwidth]{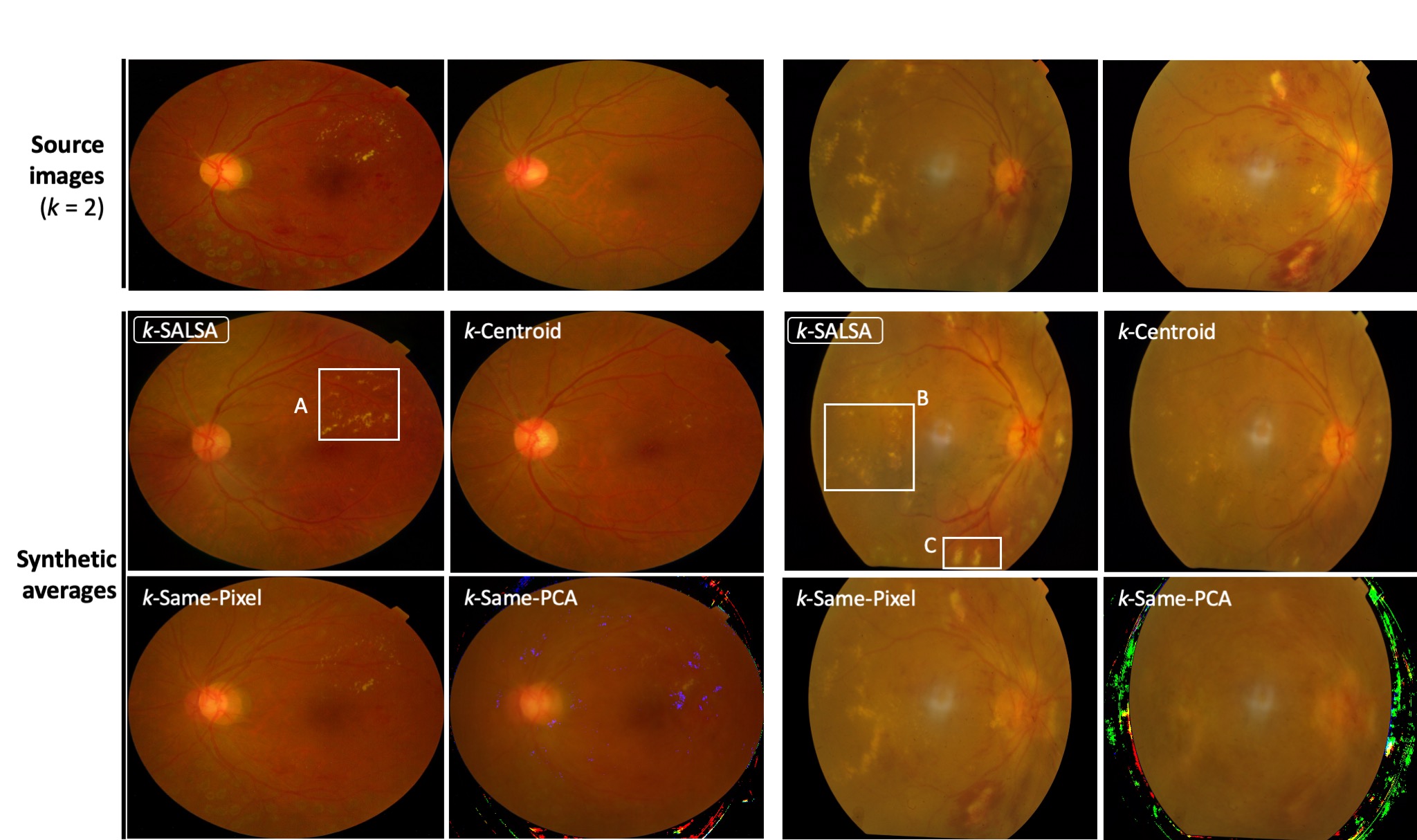}
    \caption{\textbf{Examples of synthetic average of retinal images ($k=2$).}
Two examples of $k=2$ real images (\emph{top}) along with synthetic averages generated by different methods (\emph{bottom}). $k$-SALSA better captures  disease-related features (\emph{A}, \emph{B}, \emph{C}).}
    \label{fig:k2_2}
    \vspace{-1.7em}
\end{figure*}

To complement the main results, here we include additional synthetic averages of APTOS images along with their real source images for each $k\in \{2,5,10\}$ (4 examples each): 
Supplementary Figs.~\ref{fig:k10_1}--\ref{fig:k10_4} for $k=10$;
Supplementary Figs.~\ref{fig:k5_1} and \ref{fig:k5_2} for $k=5$; and Supplementary Figs.~\ref{fig:k2_1} and \ref{fig:k2_2} for $k=2$.
The results from the EyePACS dataset are analogous.
Note that the figures for both $k=2$ and $k=5$ include two examples per figure.
Source images (\emph{top}) represent the $k$ real original images in the identified cluster, and the synthetic averages (\emph{bottom}) are generated using $k$-SALSA, $k$-Centroid, $k$-Same-Pixel, $k$-Same-PCA, respectively.
Overall, $k$-SALSA can better detect clinically relevant features in all cases.

\end{document}